\documentclass[journal,12pt,onecolumn]{IEEEtran}

\usepackage{blindtext, graphicx}
\usepackage{algorithm}

\usepackage{caption}
\usepackage{subcaption}

\usepackage{setspace}

\usepackage{bbm}

\usepackage{algpseudocode}
\usepackage{pifont}
\usepackage{tikz}
\usepackage{amsmath}
\usepackage{amssymb}
\usepackage{upgreek}

\usepackage{mathtools}

\usepackage [english]{babel}
\usepackage [autostyle, english = american]{csquotes}
\MakeOuterQuote{"}

\usetikzlibrary{arrows}

\tikzset{
	treenode/.style = {align=center, inner sep=0pt, text centered,
		font=\sffamily},
	arn_n/.style = {treenode, circle, red, font=\sffamily\bfseries, draw=black,
		text width=2em},
	arn_r/.style = {treenode, circle, red, font=\sffamily\bfseries, draw=black,
		text width=2em},
	arn_x/.style = {treenode, circle, red, font=\sffamily\bfseries, draw=black,
		text width=2em}
}

\tikzstyle{block} = [draw, fill=blue!20, rectangle, 
minimum height=3em, minimum width=6em]
\tikzstyle{sum} = [draw, fill=blue!20, circle, node distance=1cm]
\tikzstyle{input} = [coordinate]
\tikzstyle{output} = [coordinate]
\tikzstyle{pinstyle} = [pin edge={to-,thin,black}]

\begin{document}

\title{ Novel Sensor Scheduling Scheme for Intruder \\
	Tracking in Energy Efficient Sensor Networks }


\author{Raghuram Bharadwaj D.$^1$,
        Prabuchandran K.J.$^1$,
        and~Shalabh Bhatnagar$^1$
\thanks{$^1$ Authors are with the Department of Computer Science and Automation, Indian Institute of Science, Bangalore, India}
}


\maketitle


\begin{abstract}
We consider the problem of tracking an intruder using a network of wireless sensors. For tracking the intruder at each instant, the optimal number and the right configuration of sensors has to be powered.  As powering the sensors consumes energy, there is a trade off between accurately tracking the position of the intruder at each instant and the energy consumption of sensors. This problem has been formulated in the framework of Partially Observable Markov Decision Process (POMDP). Even for the state-of-the-art algorithm in the literature, the curse of dimensionality renders the problem intractable. In this paper, we formulate the Intrusion Detection (ID) problem with a suitable state-action space in the framework of POMDP and develop a Reinforcement Learning (RL) algorithm utilizing the Upper Confidence Tree Search (UCT) method to solve the ID problem. Through simulations, we show that our algorithm performs and scales well with the increasing state and action spaces. 
\end{abstract}

\section{INTRODUCTION}\label{intro}
The problem of detecting an intruder (Intrusion Detection (ID) problem) using a network of sensors arises in various applications like tracking the movement of wild animals in the forest, house/shop surveillance for safety and security and so on. In this problem, the objective of the ID system is to track one or more intruders moving in the field of a wireless sensor network (WSN).  Typically, WSNs operate on limited power supply. This imposes a limitation on the number of sensors (energy budget) that can be switched ON over a time period when tracking the intruders and thus, constraints the maximum achievable tracking accuracy.  Hence, the problem focussed in this paper is to propose a novel ID algorithm that respects such resource constraints.

We consider a variant of this problem in which an intruder is moving in a special network configuration, that is a sensor grid in which the battery operated sensors are placed in each block of the grid. At every time period, the intruder moves from one block to another according to a specified governing dynamics. The ID system decides to keep the sensors in some of these blocks ON in order to track the intruder.  Whenever the intruder moves into a block where the sensor is ON, the position of the intruder gets recorded. Instead if he moves into a block where the sensor is OFF, then the intruder's position will not be recorded for that time period. The challenge here is to decide on the optimal number and the right configuration of sensors to be powered ON in order to balance the conflicting objectives of minimizing the energy consumption of the network and maximizing the intruder tracking accuracy.


Significant body of research exists that proposes solutions to the ID problem. In \cite{survey}, a comprehensive survey of adaptive sensing methods and POMDP solution methodologies are provided.  In \cite{robust}, a sensor scheduling algorithm has been proposed for detecting an intruder with a known trajectory. In \cite{mobile}, mobile sensors along with the static sensors are deployed. They proposed an algorithm in which these mobile sensors move and detect the position of the intruder. In \cite{dsn}, a method based on the extended Kalman Filter has been applied to the directional sensor network to select a minimum set of sensor nodes. 

In \cite{sampling}, a POMDP for scheduling sensors has been formulated and Monte Carlo sampling along particle filters for belief state estimation has been proposed. This algorithm has also been applied in the context of autonomous UAV tracking \cite{miller2009pomdp}. In \cite{vv2}, the ID problem has been formulated as a partially observable Markov decision process (POMDP) and two heuristic solutions have been proposed. Note that solving the POMDP for optimal solution is computationally intractable and one often resorts to heuristic POMDP solution. In the first method, they have proposed $Q_{MDP}$ with an assumption of observation-after-control, i.e., position of the intruder at any time period will be known to the system from the immediate next and future time periods. Under this assumption, they show that the problem can be decomposed into separate sub-problems where individual decisions can be arrived at for each sensor. Each sub-problem can be solved using policy iteration \cite{sutton}. As the policy obtained is myopic in nature, \cite{vv2} developed a point-based approximation method based on the idea of Perseus \cite{pegasus}. The key idea behind this method is to find optimal action for a reachable set of simulated beliefs (see Section \ref{POMDP}) and uses the assumption that the intruder moves in a special path, i.e.,  the transition probability matrix has a special sparse structure rendering only a finite number of path configurations to be actually feasible.  
	
In many practical scenarios, the curse of dimensionality effect (exponential growth of the state and action space of POMDP) renders the problem even more challenging. Point-based techniques attempt to alleviate the curse of dimensionality effect in POMDPs by performing value function back-ups at specific belief points rather than at all the belief points. The efficiency of these techniques depends on these selected belief points. In \cite{pineau2006anytime}, an effective way of selecting the belief points has been discussed. This technique is then combined with value iteration to arrive at a Point-Based Value Iteration (PBVI) algorithm. However, this method is not scalable as it uses full-width computation accounting all possible actions, observations and the next state in the tree search \cite{uct}.
In \cite{poupart2005vdcbpi}, an algorithm VDCBPI that combines the Value Directed Compression (VDC) and Bounded Policy Iteration (BPI) techniques to tackle the state explosion problem has been proposed. Using VDC, the belief space is compressed into smaller subspace containing only the necessary information that is required to evaluate a policy. BPI is then employed to perform the policy improvement only on the reachable beliefs to determine an optimal policy. 

In \cite{maniloff2011hybrid}, a hybrid value iteration algorithm for POMDPs that combines the advantages of both the point-based techniques and the tree search methods has been proposed. In the first step of this hybrid algorithm, an offline computation is performed to obtain the upper bounds on the optimal value function. In the subsequent step, this computation is used in the online tree search for obtaining the optimal action. In \cite{prashanth2014two},  a two time-scale Q-Learning algorithm with function approximation has been developed to mitigate the curse of dimensionality problem. In their algorithm, policy gradient update is carried out on the faster timescale and Q-values update on slower time-scale.  The challenge however with function approximation primarily lies in choosing the right features for approximating the Q-values. While all the above mentioned algorithms mitigate the problem of state explosion, the problem of action explosion needs to be handled to obtain scalable solutions. 
    
In addition to the above mentioned POMDP model, many other frameworks have been considered in the literature. In \cite{active}, a region prediction sensor activation algorithm (PRSA) has been proposed where a subset of active nodes based on the position and velocity of the intruder is selected.  Among the selected subset of nodes, the lowest number of essential nodes will be switched ON. In \cite{polygon},  a model in which sensor positions are arranged in the form of a polygon has been considered and an A-star algorithm was applied for selecting the optimal nodes in the polygon. 
    
    
Our contribution in this paper is to apply general RL algorithms to solve the POMDP formulated in \cite{vv2} that tackle both the aspects of state space and action space explosion in a  novel way. Our algorithms do not make any assumptions on the structure of the network or the movement of the intruder and also do not use additional information at the controller. 

The following are our contributions:
 \begin{itemize}
 \item We solve the problem of state space explosion arising in the problem of Intrusion Detection (ID) by using the MCTS algorithm. Unlike the prior methods considered for ID problem to mitigate the state space explosion like Monte-Carlo sampling methods and the point-based value iteration methods \cite{pegasus}, the MCTS algorithm searches the belief space in a sequentially best first order. 
\item We solve the problem of action space explosion in ID problem by suitably reformulating the action space of POMDP in \cite{vv2}. This reduces the exponential number of actions to be considered in the ID problem to few finite actions. Furthermore, this makes it amenable to the application of the MCTS algorithm. 
\item Our reformulation of the action space is inspired from a simple $ID\_TG$ algorithm that greedily selects the top sensors (which implicitly ignores the current belief value). The key idea in our $ID\_\gamma\_MCTS$ algorithm is to choose the top sensors dynamically based on the current belief value. 
\item Our  $ID\_\gamma\_MCTS$ algorithm is a scalable solution to the ID problem without  any extra assumption. This renders our solution very practical. 
\end{itemize}

\section{POMDP Framework for the ID problem }{\label{POMDP}}
Let us consider a sensor grid where sensors are placed, one on each block of the grid.  The intruder moves from one block to another in the grid in each time period. The sensors that are kept ON in the grid send their observation whether the intruder was seen on that block to the central controller. Based on this information, the central controller finds the optimal action to be taken i.e., it decides how many and which sensors need to be switched ON in the next time period, and broadcasts this action to all the sensors in the network. This sequential decision making problem can be posed in the framework of Partially Observable Markov decision process (POMDP) \cite{pomdp}.

An MDP is defined via the tuple $<S,A,P,K>$. Here, $S$ is the set of states that represent the different block positions of the intruder, $A$ is the set of actions corresponding to the number and the configuration of sensors that needs to be switched ON at a given time period, $P$ is the transition probability matrix governing the state evolution where $P(s_k = i, s_{k+1}=j, u_k=a)$ represents the probability dynamics of  the intruder, i.e., probability of moving from the current block position, the state $s_k=i$ at time $k$ to the next state $s_{k+1}=j$ at $k+1$ under action $a$, and $K(s_k = i, s_{k+1}=j, u_k=a)$ corresponds to the single stage cost incurred by taking action $a$ when intruder position at time $k$ was $i$ and the position at $k+1$ is $j$. The objective here is to obtain an optimal stationary policy $\pi: S \rightarrow A$. This policy gives the optimal action to be chosen in the current block position of the intruder so that we could track the intruder in the next time period such that the long-term cost objective (constructed based on the single-stage cost function) is minimized. In order to determine the optimal stationary policy,  the ID system should know the state or position of the intruder at all the time periods. However, in our setting, whenever intruder moves to a block in which the sensor is not turned ON the intruder cannot be tracked. In this case, the position of the intruder at the next time instant will remain unknown. Thus, it is not possible to solve this problem directly in the framework of MDP.  We have to recast this problem as a partially observable MDP (POMDP) to capture the unknown state information. 

A POMDP \cite{pomdpsurvey} is formally defined as a  tuple $<S,A,Z,P,O,R>$. Here, the tuple $<S,A,P,R>$ remains the same as in the MDP setup and  $Z$ corresponds to the set of observations. As we do not directly observe the state, an observation corresponding to the state and action chosen will be obtained. This is modeled using the observation function $O$, i.e., $O:S \times A \rightarrow \pi(Z) $, where $\pi(.)$ is the probability distribution over the space of observations. In simple terms, we obtain the partial information or observation about the state in POMDP as opposed to obtaining the complete information about the state in MDP. As described earlier, in the ID problem the state information can become unknown based on our decisions and thus our problem lies in the realm of POMDP.

We now formally present the POMDP formulation for the ID problem \cite{vv2}.  
	
\begin{itemize}
\item Let $n$ denote the total number of sensors in the network.
	
\item Let the actual position of the intruder at time $k$ be denoted as $s_{k}$.
	
\item The stochastic matrix $P$ of dimension $(n+1) \times (n+1)$ models the actual transition probability of the intruder movement between various blocks. Note that the transition between states modeling the intruder movement for this POMDP does not depend on our decision or action, i.e., which sensors have been turned ON. 
	
\item Action Space : At time $k$, let $u_{k}$ denote the vector that indicates the decision on which sensors to be switched on for the next time period
	  \begin{align}
	  u_k = (u_{k,l})_{l=1,\ldots,n} \in \{0,1\}^n,
	  \end{align}
	where $0$ denotes the action to keep the sensor OFF and $1$ denotes the action to keep the sensor ON. Here, $u_{k,l}$ denotes the decision for the $l$th sensor in the $k$th time period. Note that the number of actions is exponential in the number of sensors and thus leaving us with a large number of actions from which we need to decide.
	
\item Observations: There can be three possibilities. If we track the position of the intruder, then the observation is the state of the system $s_{k}$ itself. In some cases, we may not be able to track the position of the intruder. We let $\epsilon$ to represent this situation. The final possibility is that the intruder may have moved out of the network. Let $\uptau$ indicate this situation. We assume that if the intruder moves out of the network (i.e., observation = $\uptau$ ) this information gets immediately known to the controller. So, 	
\begin{equation*}   
	o_{k+1} = \begin{cases} s_{k+1}, &  \text{if } s_{k+1} \neq \mathcal{T} ~ \text{and} ~ u_{k,s_{k+1}} = 1 \\ \varepsilon  & \text{if } s_{k+1} \neq \mathcal{T} ~ \text{and} ~ u_{k,s_{k+1}} = 0 \\ \mathcal{T}  & \text{if } s_{k+1} = \mathcal{T}. \end{cases}
\end{equation*}	

\item The history or information content available at time $k$ is: $ I_{k} =  \{o_{0},u_{0},o_{1},u_{1},.....o_{k},u_{k}\}. $	  
	  
\item The optimal action is obtained from the optimal policy according to:	  
 \begin{align}
	  u_{k} = \mu_{k}(I_{k}), 
 \end{align}
where $\mu_{k}$ denotes the optimal action selection function given the `information vector' at time $k$.
	   
\item  State Space: As the number of stages or the time slots increase, the size of history increases and thus, it becomes difficult to compute the optimal policy using finite memory. So we need to develop a different representation for the state space. We solve this problem with the help of belief vectors. The  belief vector $p$ is an $(n+1) \times 1 $ vector, in which $p_{k}(l)$ indicates the probability of the intruder being at position $l$ at time $k$. This evolves as follows: 
	 \begin{align} \label{belief}
	 p_{k+1} = &e_{\mathcal{T}} \mathbbm{1}_{\{s_{k+1} = \mathcal{T}\}} + e_{s_{k+1}} \mathbbm{1}_{\{u_{k+1,s_{k+1}} = 1\}} 
	 \\ \nonumber &+ [p_{k}P]_{\{j : u_{k+1,j} =1 \}} \mathbbm{1}_{\{u_{k+1},s_{k+1} = 0 \}},
	 \end{align}
	 
where $e_{i}$ is the vector that represents 1 at position $i$ and 0 at other positions. The notation $[V]_{S}$ represents the probability vector obtained by setting all components $V_{i}$, such that $i \in S$ to 0 and then normalizing this vector so that the sum of the components in $V$ is 1. This is to account that we did not track the position of intruder, so we are sure that intruder did not move to sensor positions which are switched on.
	 
\item We define the single stage cost model for the ID problem using two cost components. The system incurs unit cost, if we do not track the position of intruder at a given time period and 0, if we track the position. Let us define
\begin{align}
	 T(s_{k},u_{k},s_{k+1}) =  \mathbbm{1}_{\{u_{k,s_{k+1}} = 0\}}, 
\end{align}
where $\mathbbm{1}(.)$ is the indicator function. This cost $T(s_{k},u_{k},s_{k+1}) $ captures the fact whether intruder is tracked or not. The other cost component $C(s_{k},u_{k},s_{k+1})$ provides constraint on the number of sensors that can be kept awake, i.e., \begin{align}
	C(s_{k},u_{k},s_{k+1}) = \sum_{l =1}^{n} \mathbbm{1}_{\{u_{k,l} = 1\}}.
\end{align}
	
\item The long-run objective is to minimize the average expected tracking error while satisfying the limit on the number of sensors to be ON (energy budget constraints):
 \begin{align}
 \min_{u_k, k \geq 0} \lim_{m \rightarrow \infty}  \frac{\mathbb{E}[\sum_{k = 0}^{m-1} T( s_{k},u_{k}, s_{k+1})]}{m}
 \end{align}	
 subject to
 \begin{align} \label{imp}
 \lim_{m \rightarrow \infty} \frac{\mathbb{E}[\sum_{k=0}^{m-1} C(s_{k},u_{k},s_{k+1})]}{m} \leq b, 
 \end{align}
		
 where  $\mathbb{E}[\cdot]$ denotes the Expectation over state transitions and $b$ specifies the energy budget (upper bound on the number of sensors that can be kept awake) in a time period.

\item Relaxed Cost Function: We modify the single-stage cost function to include the constraint as follows:
\begin{align} \label{Rcost}
g(s_{k},u_{k},s_{k+1}) &= T(s_{k},u_{k},s_{k+1}) \nonumber \\ 
& + \lambda \times C(s_{k},u_{k},s_{k+1}), 
\end{align} 	
where $\lambda \in [0,1]$ is a suitable threshold for tracking error and budget constraints. If we set $\lambda$ to 0, it is similar to having unlimited budget, in which case all the sensors will be kept ON all the time. On the other hand, $\lambda = 1$ implies there is strict budget constraint, in which case, all the sensors will be switched OFF. In this way, by optimally tuning $\lambda$, the algorithm could support different levels of tracking error and average energy budget.
\end{itemize}

Now we have modeled the ID problem using the POMDP framework. The uncertainty in the state can be removed by treating belief vector as our new state. We now have all the ingredients for an MDP and one would hope to utilize the standard dynamic programming methods like policy iteration and value iteration \cite{vol1} for solving the MDP. However, these methods operate on finite state space whereas the possible belief vector, which forms the new state space for this problem is uncountably infinite. Therefore, finding an optimal policy for this problem by solving the MDP is intractable \cite{vv} leading us to focus on developing a suitable algorithm that can handle a large state and action space. This would ensure good tracking performance while meeting the energy budget constraints.

\section{Our ID Algorithms}{\label{algo}}
In this section, we describe our RL algorithms.
%

\subsection{Greedy Algorithm ($ID\_TG$)}
This is a simple greedy algorithm in which the top probable sensor positions the intruder might move to will be turned ON for tracking during each time period. The number of sensors that will be kept ON is decided based on a chosen parameter $\gamma$, where $\gamma \in  (0,1)$. The first step of this algorithm involves obtaining the approximate belief vector for the next time period. Note the belief vector gives the probability of the intruder reaching various possible positions in the next time period. In order to compute the belief vector, we use \eqref{belief}. Then we select those positions in the belief vector that sum up to $\gamma$ starting from the highest probability value. 

We could view $\gamma$ as a minimum confidence index on tracking the position of intruder. For example, consider the approximate belief vector for the next time period to be  $ [ 0.3, ~0.3, ~ 0.2, ~0.2]$. Let $\gamma = 0.6$. Then the sensors at positions 1 and 2 will be switched ON for the next time period and with confidence of 0.6, we know that intruder will be present in either position 1 or 2. 

We can see that this algorithm is very easy to implement and at the same time solves the problem of state and action space explosions. However, at every instance, $\gamma$ chosen is constant and independent of the number of non-zero values in belief vector. Consider a scenario in which the belief vector is `dense'. By `dense', we mean that there are many non-zero probability positions in the belief vector whose values are very close to each other. Suppose we did not track the position of intruder in that time slot. Then, the belief vector grows further dense (see \eqref{belief}). In this case, we would like to use a higher $\gamma$ so that we could have more number of sensors ON and track the position of the intruder. Otherwise, the belief vector would grow more dense resulting in bad tracking accuracy. On the other hand, if the belief vector is sparse, then even a smaller  $\gamma$ would give better tracking. Also, we do not know, the best choice of $\gamma$ for a given budget and the current belief. In summary, having a constant gamma value at every time period is not the right choice for optimal performance. However, the idea of using  $\gamma$ as a handle for deciding the actions instead of directly searching over the number and configurations of sensors will play a crucial role in solving the action space explosion problem in our proposed algorithm $ID\_\gamma\_MCTS$.

\begin{algorithm}
\caption{$ID\_TG$}
\label{greedy}
\begin{algorithmic}[1]
	\State $n$ $\leftarrow$ Number of sensors in the network.
	\State $\gamma$ $\leftarrow$ predefined value entered by the user. 
	\State $k=0$, $p_{0}$ $\leftarrow$ Initial belief vector, 
	\Procedure{$ID\_POMDP$}{}
	\While{Intruder is in the Network}
	\State $ABV_{k+1}$ = $p_{k}\times P$
	\State $g = 0 ; u(k,l) = 0,\hspace{0.1cm} \forall l = 1...n$
	\While{$g$ $\leq \gamma$}
	\State $l \leftarrow  ABV_{l} $ with maximum probability. 
	\State $u(k,l) = 1$, $g$ := $g$ + ABV(i), // Sensor $k$ will be switched on.
	\State Remove position $l$ from ABV. 
	\EndWhile
	\State Update Belief as in \eqref{belief}
	\State $ k+= 1$
	\EndWhile
	\EndProcedure
\end{algorithmic}
\end{algorithm}

\subsection{Monte Carlo Tree Search (MCTS)}
In this section for completeness, we first describe the idea of MCTS  for determining the optimal decisions for an MDP and then we adapt this algorithm to the POMDP setting. For more details about the algorithm the reader is referred to \cite{uct} . MCTS has gathered a lot of attention in recent times due to its success in playing strategic games like GO \cite{nature}. The main idea behind this algorithm is to run multiple simulations (with the help of simulator placed at  the controller) from the current state to determine the best action in each iteration. We begin with a single node where the node represents the current state. Then, we select an action for the given state using the Upper Confidence Bound for Trees (UCT) rule. The general UCT rule is given in \cite{uct} as follows:

\begin{align}\label{uctf}
UCT\_action  =\arg \max_{j} {\hat{X}_{j} + C * \sqrt{\log N / N_{j}}},
\end{align}

where $\hat{X}_{j}$ denotes the estimate of the average/discounted long-run reward (negative of the long-term cost) one obtains by selecting action $j$ starting from the current state, $N$ denotes the total number of runs so far and $N_{j}$ corresponds to the total number of runs by selecting action $j$.

In the beginning, all actions have 0 count, i.e., $N_{j} = 0 $. Therefore, we can see that an unexplored action will have a higher probability of being selected on each simulation run. After all the actions have been explored, the action that has led to the highest reward collected, i.e, the term $\hat{X}_{j}$ gets selected. In this manner it balances both exploration and exploitation. When an action is selected, we obtain the single stage reward/cost and a next state. Then, a new node and a new edge is added to the tree, where the edge represents the action we have chosen and the new node represents the next state obtained. We continue this construction, adding new nodes and edges to the tree till the desired depth. In this manner, we obtain a Monte-Carlo sample trajectory and a sample for the  estimate of the long-run reward/cost.  We could obtain multiple trajectories by trying out different actions (including the tried action) in the current state for getting better estimates of the long-run reward. This procedure is now repeated again from the root node till the timeout \cite{uct}. At the end of the timeout, we pick the action that has maximum long-run reward.  This method helps in breaking the curse of dimensionality of the state space by sampling the state transitions instead of considering all possible state transitions to estimate long-run reward. If the exploration is done in an optimal manner, \cite{uct} has shown that the algorithm converges to the optimal policy.
 
 \subsubsection{$ID\_MCTS$ Algorithm}
In this algorithm, we run the MCTS \cite{uct} on our setting. That is, for the belief vector at time $k$, we run the MCTS algorithm and obtain the action to be executed for the next time instant using the UCT action selection \eqref{uctf}. At each iteration of the algorithm, the position of the intruder needs to be known for the simulator to generate the next state. We can estimate it from the belief vector in two ways. The first method involves sampling a position from the belief vector according to its distribution at each iteration. The second approach selects the position with maximum probability at every iteration. The second approach is very natural and we employ it in our algorithm.
 
From the experiments, we observe that the algorithm increases the number of sensors that are kept ON in the subsequent intervals if it does not detect the position of the intruder for a large number of contiguous time periods. Similarly, as the intruder gets tracked continuously, the number of sensors that are kept ON in the subsequent time periods reduces.
 
As discussed earlier, this algorithm solves the problem of state explosion. However, the algorithm requires all the (exponential) number of actions (the different configuration of sensors out of $n$ block positions) to be tried out sufficient number of times before the timeout. There can be $2^{n}$ actions as $n$ is the total number of sensor position actions possible. As the size of network increases, the action space exponentially increases and it takes large amounts of time to execute all the actions due to which some actions  might not be tried. Thus, the action we finally obtain at the end of the timeout will be `sub-optimal'.  We overcome the main difficulty due to large action space by appropriately changing the action space (all possible sensor positions) of the POMDP to a discretized parameter $\gamma \in [0,1]$  that was fixed in the $ID\_TG$ algorithm and develop the final algorithm $ID\_\gamma\_MCTS$ that solves the problems of both state and action explosions.  
 
\subsection{$ID\_\gamma\_MCTS$ Algorithm}
In the $ID\_ \gamma\_MCTS$ algorithm, we let the action space to be a predefined set of $\gamma$ values in the range 0 to 1 (with a uniform gap of 0.05). Note that the number of possible actions for this algorithm is constant unlike the exponential actions in the earlier algorithms in the literature. We run MCTS to obtain the optimal $ \gamma$ value for the present belief vector. As we have discretized the action space, all the actions are chosen a sufficient number of times. The idea behind this construction is that it suffices to know how many sensors need to be kept awake in terms of $\gamma$ during each time period instead of explicitly knowing the exact configuration of the sensors. We could then use this value of $\gamma$ to select the top probable positions in the belief vectors as described in Algorithm \ref{greedy}. These sensor positions will be kept ON for the next time period. This process is repeated until the intruder moves out of the network. The complete $ID\_ \gamma \_MCTS$ algorithm is described in Algorithm \ref{Algorithm3}.

As described above, Algorithm 3 imbibes the ideas of both $ID\_TG$ and $ID\_MCTS$ and totally solves the problems of state and action explosions. Moreover, we observe that the $\gamma$ value at each time period is selected dynamically and is not kept constant, which is a significant drawback for the $ID\_TG$ algorithm. 

From the experiments, we observe that in few cases the actual path of the intruder and the path that we are estimating diverge. This happens whenever the belief vector gets too dense and $\gamma$ selected is not very high. Then, we lose the position of the intruder and in this case, the belief vector grows more dense in the subsequent time periods. This results in the intruder not getting tracked for many contiguous time intervals and which affects the tracking accuracy. To overcome this problem, we introduce the restart mechanism. This involves switching on all the sensor positions that have non-zero probability whenever the belief vector has more than the threshold number of non-zero values. We observe this divergence problem occurs rarely and thus the restart cost can be practically ignored.

\begin{algorithm}
\caption{$ID\_MCTS$}
\label{Algorithm2}
\begin{algorithmic}[1]
	
   \State Action\_Space - set of all $2^{n}$ configurations of sensors. 	
   \State $get\_action()$ - function to compute the optimal action for a given belief. 
   \State $next\_position()$ : function to compute the next actual position of the intruder. 
   \State $get\_cost()$ : It computes the single stage cost as described in \eqref{Rcost}.	
   \State $k = 0$, $tot\_cost = 0$
   	
	\Procedure{MDP}{}
	\While{Object is within the Network}
	
	\State $a_{k} = get\_action(p_{k})$ 
	\State $s_{k+1} = next\_position(s_{k})$
	
	\State $p_{k+1} \leftarrow $ Update the belief vector as in \eqref{belief}
	
	\State tot\_cost $+= get\_cost(s_{k},a_{k},s_{k+1})$, $k += 1$
		
	\EndWhile
	
	\EndProcedure

	\Procedure{$get\_action$}{p}
	\While{$Time\_Out  \geq$ 0}
	
	\State s = $\arg\max_{s} p(s)$ 
	
   \State $MCTS(s,p,1)$
	
	\EndWhile
	\State return $\arg\min_{a} \hat{X} {(p,a)}$;
	
	\EndProcedure
	\Procedure{MCTS}{$s,p,depth$}
	\State if (depth $\geq$ MAXDEPTH ) return 0;
	
	\State Select an action $a$ using the $UCT$ policy as in \eqref{uctf} (or any other action exploration strategies)     		
	\State  $ s\_new \sim Simulator(s)$ 
	\State $p\_new$ $\leftarrow$ Update belief as in \eqref{belief}
	\State  $Cost \leftarrow get\_cost(s,a,s\_new) + $
	\State $\alpha * MCTS(s\_new,p\_new,depth +1)$	
	\State  $\hat{X}{(p,a)} += (Cost - \hat{X}{(p,a)}) / N(p,a)$
	\State $ N(p,a) += 1 $
	
	\EndProcedure	
\end{algorithmic}
\end{algorithm}

\begin{algorithm}
\caption{$ID\_\gamma\_MCTS$}
\label{Algorithm3}
\begin{algorithmic}[1]
	\State Action\_Space : Discrete set of gamma values entered by user	
	\State $get\_gamma()$ - computes the best gamma value for a given belief
	\State $get\_action()$ - sets the action for the given belief and $\gamma$ value	
	\State $k = 0$, $tot\_cost = 0$	
	\State $s_{0} \leftarrow$ Initial position
	\State $p_{0} \leftarrow$ Initial belief
	\Procedure{MDP}{}
	
	\While{Object is within the Network}
	
	\State $ \gamma = get\_gamma(p_{k})$
	
	\State $ a_{k} = get\_action(p_{k},\gamma)$
	\State $s_{k+1} = next\_Position(s_{k})$
	
	\State $p_{k+1} \leftarrow$ Update belief as in \eqref{belief}
	
	\State $tot\_cost += get\_cost(s_{k},a_{k},s_{k+1})$
	
	\EndWhile
	\EndProcedure

\end{algorithmic}
\end{algorithm}

\section{Experiments and Results}{\label{exp}}

We considered three different configurations of the sensor network. In our first setting, we considered a 1-dimensional sensor network with 41 sensors with probability transition matrix similar to the one considered in \cite{vv2}. At the start of the experiment, the intruder is placed at the center of the network with the movement constraint that he could either move 3 positions left or 3 positions right from the current position. The restart threshold for this setting is set to 14. In the second setting, we ran our algorithms on a two-dimensional sensor grid of dimension $8\times 8$. The feasible movements of the intruder are left, right, down, up and along all the diagonals. In our final experiment,  we ran our algorithms on 2-D grid with dimension $16\times 16$.  In both the second and third settings, we generated the transition probability matrix randomly and the restart threshold is set to 20.

We averaged the results of our simulations over 30 time periods. The MCTS algorithm was run for 500 iterations in every time period. We computed the long-term cost in the experiments as the discounted sum of single-stage costs for learning the decisions. Note that by choosing the discount factor close to 1, the actions learned for the long-run discounted cost setting will be similar to the actions learned for the long-run average cost setting. We have set the discount factor $\alpha = 0.9$. For the $ID\_\gamma\_MCTS$ algorithm, we chose the $\gamma$ parameter corresponding to the actions by discretizing the interval $[0,1]$ in steps of 0.05. Thus, there are totally 20 discretized actions in the action space. The $Q_{MDP}$ method is implemented by solving the policy iteration in \cite{vv2} and making the assumption of observation-after-control, i.e., $p_{k+1} = e_{b_{k+1}}$ at the controller. In the plots, the X-axis corresponds to the average number of sensors awake which is computed as the ratio of the total number of sensors switched ON during the run of the algorithm and the total number of time periods and the Y-axis corresponds to the average tracking error which is obtained as the ratio of the number of time periods in which the intruder is not tracked and the total number of time periods.

\begin{table}[h]
\centering
\caption{$\lambda$-values and corresponding results for $ID\_\gamma\_MCTS$ algorithm on 41-sensor network}
\label{table 1}
\begin{tabular}{|l|l|l|l|l|l|l|l|l|l|l|l|}
\hline
\textbf{$\lambda$ - Value}    & 0 & 0.1     & 0.15    & 0.2     & 0.3  & 0.35    & 0.4     & 0.5     & 0.6     & 0.8     & 1                      \\ \hline
\textbf{Avg. Sensors awake}  & 6 & 4.17  & 3.78   & 3.57 & 3    & 2.73 & 2.46 & 1.83  & 1.34 & 0.77 & 0                      \\ \hline
\textbf{Avg. Tracking Error} & 0 & 0.16 & 0.20 & 0.23 & 0.32 & 0.41 & 0.40 & 0.44 & 0.48 & 0.53 & \multicolumn{1}{r|}{1} \\ \hline
\end{tabular}
\end{table}

\begin{table}[h]
\centering
\caption{$\lambda$-values and corresponding results for $ID\_\gamma\_MCTS$ algorithm on 2D network of size 8$\times$ 8}
\label{table 2}
\begin{tabular}{|l|l|l|l|l|l|l|l|l|l|l|l|}
\hline
\textbf{$\lambda$ - Values}    & 0 & 0.1  & 0.15 & 0.2  & 0.3  & 0.35 & 0.4  & 0.5  & 0.6  & 0.8  & 1                      \\ \hline
\textbf{Avg. Sensors awake}  & 9 & 7.32 & 6.84 & 5.90 & 4.40 & 3.75 & 3.05 & 1.94 & 1.34 & 0.84 & 0                      \\ \hline
\textbf{Avg. Tracking Error} & 0 & 0.14 & 0.19 & 0.22 & 0.31 & 0.40 & 0.50 & 0.49 & 0.54 & 0.56 & \multicolumn{1}{r|}{1} \\ \hline
\end{tabular}
\end{table}

\begin{table}[h]
\centering
\caption{$\lambda$-values and corresponding results for $ID\_\gamma\_MCTS$ algorithm on 2D network of size \hspace{1 in}16 $\times$ 16}
\label{table 3}
\begin{tabular}{|l|l|l|l|l|l|l|l|l|l|l|l|}
\hline
\textbf{$\lambda$ - Value}    & 0 & 0.1  & 0.15 & 0.2  & 0.3  & 0.35 & 0.4  & 0.5  & 0.6  & 0.8  & 1                      \\ \hline
\textbf{Avg. Sensors awake}  & 9 & 6.89 & 6.68 & 6.06 & 4.77 & 3.86 & 3.39 & 2.02 & 1.28 & 0.69 & 0                      \\ \hline
\textbf{Avg. Tracking Error} & 0 & 0.13 & 0.21 & 0.24 & 0.37 & 0.42 & 0.46 & 0.55 & 0.54 & 0.55 & \multicolumn{1}{r|}{1} \\ \hline
\end{tabular}
\end{table}

In our experiments, we run our algorithms for different values of $\lambda$ and choose the $\lambda$ value that meets the given budget constraints (the average number of sensors awake) and also has the lowest tracking error. We plot the average number of sensors awake and its corresponding tracking error. The $\lambda$ value for which we obtained results for our $ID\_\gamma\_MCTS$ on all the 3 settings is shown in Table \eqref{table 1},\eqref{table 2},\eqref{table 3}. Note that for different algorithms different values of $\lambda$ could lead to the same average number of sensors.  As the size of the sensor network is large in the second and third settings, our $ID\_MCTS$ algorithm doesn't scale up owing to the action space explosion. Thus, we do not show results for $ID\_MCTS$ algorithm in the second and third plots and compare mainly our $ID\_\gamma\_MCTS$ algorithm against $Q_{MDP}$ given in \cite{vv2}. All the results for $ID\_\gamma\_MCTS$ algorithm are averaged across 10 simulation runs of the experiment.

\begin{figure}[h!]
\begin{center}
 \includegraphics[width = 8cm,height=5cm]{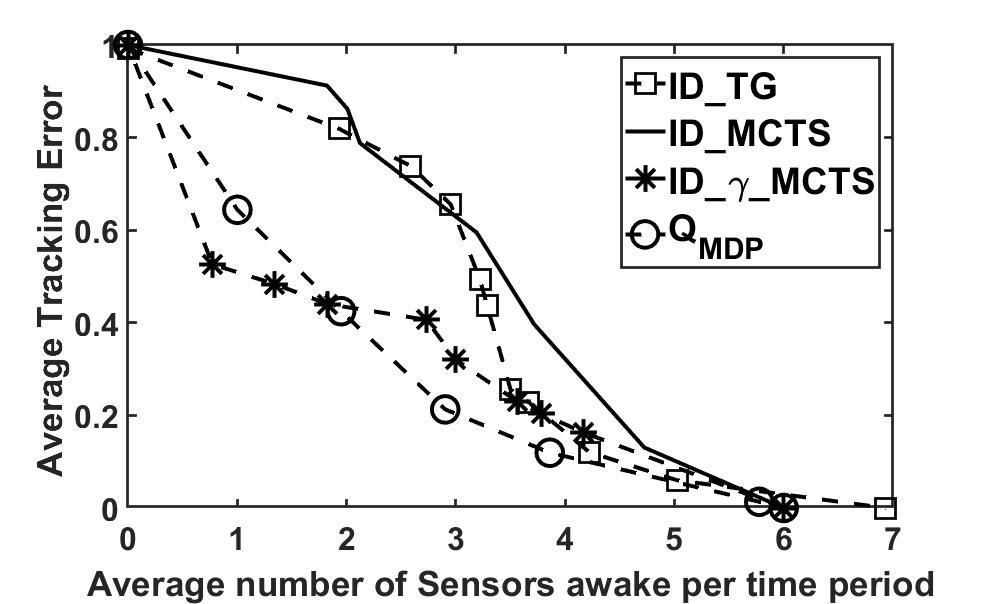}
 \caption{1D Sensor Network with 41 sensors}
 \end{center}
 \end{figure}
 
 \begin{figure}[h!]
\begin{center}
 \includegraphics[width = 8cm,height= 5cm]{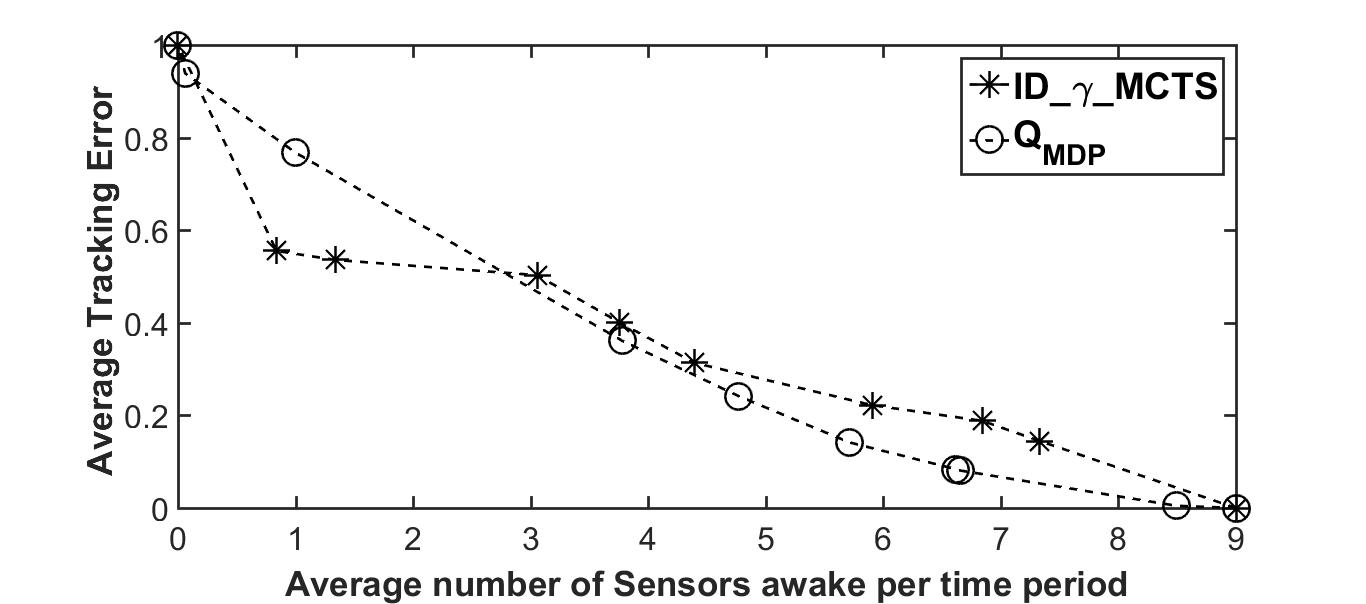}
 \caption{2D Sensor Network of size $8\times8$}
 \end{center}
 \end{figure}
 
 \begin{figure}[h!]
\begin{center}
 \includegraphics[width = 8cm,height= 5cm]{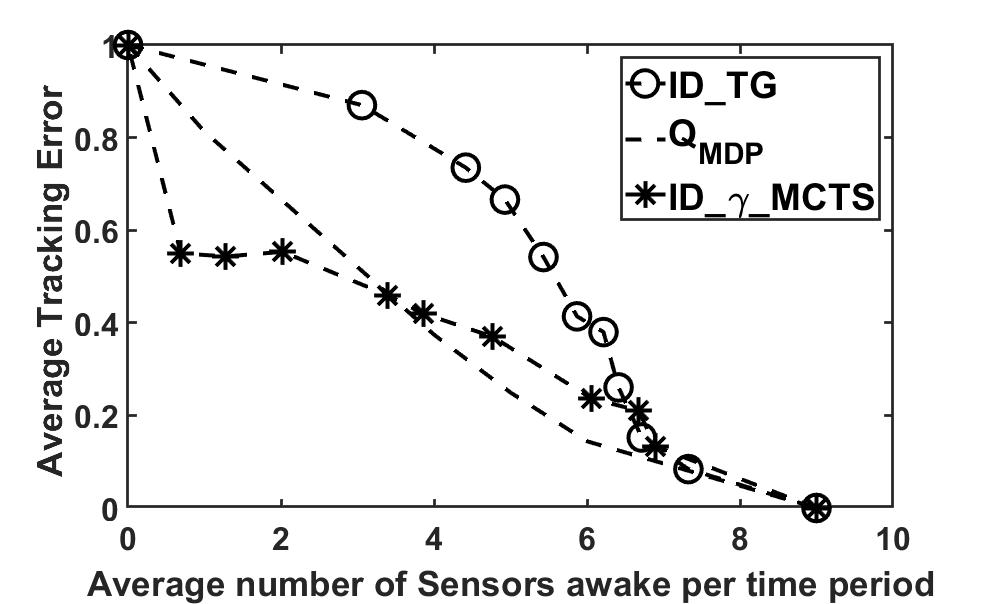}
 \caption{2D sensor Network of size $16\times16$}
 \end{center}
 \end{figure}


In the Figure 2, we observe that $ID\_\gamma\_MCTS$ outperforms $ID\_MCTS$. This is because the cardinality of the action space in $ID\_MCTS$ is 127 where as in $ID\_\gamma\_MCTS$ the cardinality is 20. However, we can see that $Q_{MDP}$ performs better than both our proposed algorithms. This is mainly because of the observation after control assumption (that we do not impose in our setting) and partly due to the small size of the sensor network. As noted earlier, this assumption imposes a severe constraint and we don't make this assumption. Thus, it is noteworthy that our algorithm performs well even without this assumption. In the Figure 3, we observe that the performance of $ID\_\gamma\_MCTS$  in comparison with $Q_{MDP}$ has improved over a 2D sensor network. In fact, it performs better than $Q_{MDP}$ until the average number of sensors is 3. In the Figure 4, we see the performance of the $ID\_\gamma\_MCTS$ algorithm has further improved. To conclude, the $ID\_\gamma\_MCTS$ performs better than other algorithms when the sensor network become very large which is typical in most of the practical applications.

As discussed earlier, since $ID\_TG$, uses a constant $\gamma$ its performance is poor (see Figs. 2 and 4). In summary, we can conclude that $ID\_\gamma\_MCTS$ without making any strong assumptions performs and scales well.



\section{Conclusion}
In this work, we proposed three algorithms for solving the problem of intruder detection under energy budget constraints. Our first algorithm $ID\_ TG$ is greedy, simple and easy to implement. However, due to its static nature, it does not yield good performance. Our second ($ID\_MCTS$) algorithm is a suitable adaptation of the MCTS algorithm for the ID problem. This solves the problem of state space explosion problem, however, the action space is still large and remains to be handled. Our final algorithm ($ID\_\gamma\_ MCTS$) combines ideas from our earlier two algorithms and totally solves both the state and action space explosion problem and this algorithm is the first of its kind for the ID problem. In our MCTS based algorithms, as we only need to account for the belief vectors that are encountered during the run, which are finite, holds the key to mitigating the state space explosion. Further, as we suitably modify the action space using predefined set of $\gamma$ values the action space explosion is also effectively handled. From simulations, we observed that the $ID\_\gamma\_MCTS$ algorithm is our best algorithm when the network size becomes large and performs well without making any assumption on the intruder movement or the intruder position information. In the future, we would like to extend this problem when multiple intruders are moving in the sensor network. In this scenario, we could consider the possibility of more than one intruder moving in the network with possibly different transition probabilities and their movements could be correlated as well. Our objective would be to track the positions of as many intruders as possible satisfying energy budget constraints. 

\section*{Acknowledgment}

The above work is supported by Joint DRDO-IISc Programme to Advance The Frontiers of Communications,
Control, Signal Processing and Computation.


 \bibliographystyle{IEEEtran}
 \bibliography{IEEEabrv,reference}

\end{document}